\documentclass[acmsmall,nonacm]{acmart}
\usepackage{tcolorbox}
\usepackage{array}
\usepackage{hyperref}
\usepackage{enumitem}

\AtBeginDocument{%
  }

\setcopyright{acmlicensed}
\copyrightyear{2025}
\acmYear{2025}

\begin{document}



\title{World Models in Artificial Intelligence: Sensing, Learning, and Reasoning Like a Child}
\subtitle{\textit{Reasoning in Artificial Intelligence depends not on larger models, but on teaching them to think}}

\author{Javier Del Ser}
\orcid{0000-0002-1260-9775}
\affiliation{%
  \institution{TECNALIA, Basque Research \& Technology Alliance (BRTA)}
  \city{Derio}
  \country{Spain}
}
\affiliation{%
  \institution{Department of Mathematics, University of the Basque Country (UPV/EHU)}
  \city{Leioa}
  \country{Spain}
}
\email{javier.delser@tecnalia.com}

\author{Jesus L. Lobo}
\orcid{0000-0002-6283-5148}
\affiliation{%
  \institution{TECNALIA, Basque Research \& Technology Alliance (BRTA)}
  \city{Derio}
  \country{Spain}
}
\email{jesus.lopez@tecnalia.com}

\author{Heimo M\"uller}
\email{heimo.mueller@medunigraz.at}
\orcid{0000-0002-9691-4872}
\affiliation{%
  \institution{Medical University Graz}
  \city{Graz}
  \country{Austria}
}

\author{Andreas Holzinger}
\orcid{0000-0002-6786-5194}
\affiliation{%
  \institution{University of Natural Resources and Life Sciences Vienna}
  \city{Vienna}
  \country{Austria}
}
\email{andreas.holzinger@boku.ac.at}

\renewcommand{\shortauthors}{Del Ser, L. Lobo, M\"uller \& Holzinger}

\begin{abstract}
World Models help Artificial Intelligence (AI) predict outcomes, reason about its environment, and guide decision-making. While widely used in reinforcement learning, they lack the structured, adaptive representations that even young children intuitively develop. Advancing beyond pattern recognition requires dynamic, interpretable frameworks inspired by Piaget’s cognitive development theory. We highlight six key research areas -- physics-informed learning, neurosymbolic learning, continual learning, causal inference, human-in-the-loop AI, and responsible AI -- as essential for enabling true reasoning in AI. By integrating statistical learning with advances in these areas, AI can evolve from pattern recognition to genuine understanding, adaptation and reasoning capabilities. 
\begin{tcolorbox}
\textbf{KEY INSIGHTS:}
\begin{itemize}[leftmargin=*]
\item AI needs structured World Models to move beyond pattern recognition and achieve true reasoning. Unlike humans, who develop intuitive causal and physical understanding from experience, AI remains limited to statistical correlations. 

\item Cognitive development principles, inspired by Piaget’s theory, offer a way forward by guiding AI through structured learning stages, helping it actively build knowledge rather than memorizing patterns. 

\item Key research areas like physics-informed learning, neurosymbolic AI, open-world machine learning, causal inference, human-in-the-loop AI, and responsible AI can enable AI to reason, generalize, and interact meaningfully with the world.
\end{itemize}
\end{tcolorbox}
\end{abstract}



\maketitle

\section{Can Modern AI Systems reason? Are we already there?}

Statistical machine learning and Artificial Intelligence, particularly with the advent of Large Language Models (LLMs), have seen remarkable success in recent years \cite{BengioLeCunHinton:2021:DeepLearningAI}. These generative models excel at natural language processing tasks and have expanded into multiple data modalities by leveraging massive datasets and billions of parameters. Their achievements have fueled enthusiasm, investment, and innovation, enabling transformative applications across various fields \cite{lobo2024can}.  

However, despite their impressive capabilities, these models have fundamental limitations. While they can generate human-like language and behavior by capturing statistical correlations, they lack essential cognitive abilities such as abstract reasoning, contextual understanding, causal understanding, and nuanced judgment. As Shanahan \cite{Shanahan:2024:LLMs} notes, even the most advanced AI systems remain fundamentally different from humans, particularly in their lack of common sense \cite{DavisMarcus:2015:commonSense}. This gap has heightened concerns about the safety of modern AI regarding its alignment with human values and design objectives \cite{bengio2024managing}, which call for the urgent development of responsible AI frameworks \cite{Shneiderman:2021:ResponsibleAI}.  

A useful way to understand this limitation is by examining how young children interact with and learn from their environment. Human learning is deeply rooted in sensory experiences—seeing, hearing, touching, tasting, and smelling—during interactions with the physical world \cite{Piaget:1954:RealityWorld}. Even from an early age, children can infer causal relationships from only a few observations \cite{GriffithsEtAl:2011:BayesBlickets,TenenbaumEtAl:2011:GrowMind}, an ability that reflects the paradox famously described by Moravec: ``What is easy for computers is difficult for humans, and what is easy for humans is difficult for computers'' \cite{Moravec:1988:MindChildren}.  

\begin{tcolorbox}
Jean Piaget’s seminal work on child development \cite{Piaget:1963:WorldModel} offers profound insights into how humans perceive, interpret, and reason about the world. His theory of cognitive development, particularly the principle of \emph{constructivism}, asserts that knowledge is not passively absorbed but actively constructed through experience and interaction. Piaget outlined four stages of cognitive development: i) the \emph{sensorimotor stage} (0–2 years), where learning occurs through direct sensory experience and actions; ii) the \emph{preoperational stage} (3–7 years), where symbolic thinking emerges, though logical reasoning remains limited; iii) the \emph{concrete operational stage} (8–11 years), where logical reasoning develops but is still grounded in concrete experiences; and iv) the \emph{formal operational stage} (12+ years), where abstract and hypothetical thinking becomes possible. His theory underscores that cognitive development is a dynamic process driven by biological maturation and interaction with the environment, emphasizing the role of active learning.  
\end{tcolorbox}

Humans naturally develop an intuitive understanding of physics, using it to navigate daily life, predict how objects will behave, and infer causal relationships from experience. Even young children construct causal models to make sense of the physical world \cite{GopnikEtAl:2004:CausalLearning}. This experiential knowledge enables humans to adapt to new environments, anticipate future outcomes, and distinguish between what is plausible, beneficial, or impossible, which are key components of \emph{common sense} reasoning.  

Given AI's limitations, Piaget’s constructivist approach prompts an important question: \textbf{Can we draw on constructivist principles to go beyond pattern recognition and develop AI systems capable of understanding and reasoning about the world?} This paper explores this question highlighting key areas in AI where constructivist principles can be integrated into AI to actively build structured, interpretable, and reliable World Models. By enabling AI to acquire and refine knowledge through interaction and experience, much like humans do, we outline a path toward more robust, adaptable, and intelligent AI systems.

\section{What is Reasoning in AI? Why is modern AI unable to understand and reason? How could World Models surpass these limitations?}

Reasoning in Artificial Intelligence allows systems to make predictions by integrating input data, prior knowledge, and learned patterns. Unlike simple pattern recognition, which merely identifies correlations, reasoning requires synthesizing information, managing uncertainty, and drawing meaningful connections between inputs and decisions. AI reasoning generally falls into three interrelated forms. In some cases, it follows a structured, rule-based approach, applying general principles to specific situations, much like a self-driving car recognizing a red light and stopping in accordance with traffic laws. At other times, it operates inductively, identifying patterns from past observations to infer broader rules, as seen in a weather model that associates dark clouds with impending rain. AI can also reason abductively, forming the most plausible explanation when faced with incomplete data, similar to how a medical system, when presented with symptoms like fever and fatigue, hypothesizes influenza as the likely cause. Each of these reasoning processes help AI trascend simple correlation-based pattern modeling toward deeper understanding.

True reasoning requires AI to process information while grasping relationships, causal links, and contextual subtleties—key to human-like intelligence. Yet, we are far from this goal. Studies show that modern AI, particularly LLMs, often lacks genuine causal reasoning \cite{zevcevic2023causal,shi2023large,mirzadeh2025gsmsymbolic}. Though these models generate fluent, contextually appropriate text, they struggle with causal inference, multi-step logic, and cross-domain integration. Illusory reasoning can arise from seamless short-term predictions, as LLMs optimize for likely next words rather than true understanding. When faced with reasoning tasks -- such as solving math problems, playing chess, or interpreting ambiguity -- they may confidently produce incorrect answers. Their fluency masks an inherent brittleness, highlighting a core limitation: without a structured cognitive framework, modern AI remains confined to pattern recognition rather than genuine reasoning.

The reasoning limitations of large models stem from their agglomerative construction. These models aggregate vast datasets, learning patterns through an optimization objective that prioritizes predicting accurately the next word, phrase, token, or visual patch. While effective at capturing statistical correlations, this approach lacks the structured logical rules and causal frameworks essential for genuine reasoning. Post-training refinements, such as alignment strategies or reinforcement learning from human feedback, improve outputs but do not address their core deficiency. As a result, large models may superficially simulate reasoning but cannot replicate human cognition, as their architecture prioritizes pattern discovery over logical understanding.

A crucial shortfall of modern AI is the lack of \textbf{World Models} \cite{ha2018world}. Unlike humans, who construct internal representations to reason, predict, and decide, large models rely solely on statistical associations in their training data. A World Model enables AI to develop a structured, dynamic understanding of its environment, capturing relationships, rules, and causal links. With such a model, AI can reason about cause and effect, simulate future outcomes, and refine understanding through real-world interactions. Essentially, a World Model acts as an internal cognitive map, allowing AI to evolve toward structured, context-aware decision-making. It provides the necessary framework for integrating perception, representation, reasoning, and generalization, forming the foundation for more advanced, human-like intelligence. 

To provide genuine reasoning capabilities, a World Model must possess several key properties. It must first include \emph{structured, causal representations} of the physical world, allowing AI to reason beyond patterns and address causal inference in decision-making. Additionally, the model should support the \emph{identification of concepts}, enabling the AI to abstractly manipulate them and construct logic, similar to how humans reason. Finally, a World Model should be \emph{interactive, adaptable and teachable}, so that its understanding can be refined through human feedback, real-world experiences and expert demonstrations, just like a parent explains the complexities of the world to a child. An intelligent system having these properties could ensure that its reasoning aligns with ethical and domain-specific constraints, making it more robust, reliable, and respectful with human values. 

As we will now argue in detail, these properties can be better guaranteed if World Models are constructed in alignment with Piaget's theory of cognitive development. Piaget’s emphasis on learning through interaction with the environment, active construction of knowledge, and the dynamic reorganization of cognitive structures offers a compelling framework for developing World Models that truly reason in ways analogous to human cognition.

\section{Towards a Constructivist Theory of World Models}

Jean Piaget’s theory of cognitive development \cite{Piaget:1954:RealityWorld} offers a paradigmatic strategy for overcoming the reasoning limitations of modern AI systems by modeling how humans progressively build an understanding of the world. Piaget emphasized the importance of \emph{schemas}, namely, mental frameworks for organizing knowledge, and formalized the processes of assimilation and accommodation of newly acquired knowledge, through which humans refine their understanding of new experiences. Mimicking these developmental stages in machine learning could allow AI systems to construct dynamic World Models that evolve over time, enabling them to grasp physical laws, temporal dynamics, and complex causal relationships. 

The fields of developmental psychology and Artificial Intelligence offer distinct but complementary perspectives on how systems (whether biological or artificial) learn, adapt, and structure knowledge, capabilities that are essential to reasoning itself. Piaget’s theory of cognitive development, which delineates how children acquire and refine their understanding of the world, aligns conceptually with the principles underlying World Models in AI. We next provide insights into how these frameworks converge in their approach to representing knowledge and adapting to the environment, and which aspects are not still met by current AI methods.

\noindent\paragraph{\textbf{Cognitive Schema and World Models}} Piaget’s theory \cite{Piaget:1954:RealityWorld} posits that \emph{schemas} are the building blocks of cognitive development. They are mental frameworks that children use to interpret experiences. These schemas evolve as new information is integrated. For instance, a child’s schema for ``dog'' may include barking and four legs, but encountering a cat triggers the refinement of this schema. A schema encompasses both a knowledge category and the process of acquiring it, adapting through \emph{assimilation} (integrating new experiences) and \emph{accommodation} (modifying frameworks to incorporate new information). Similarly, in AI, World Models should provide structured understandings of an environment, predicting outcomes and guiding decisions. Neural networks, for instance, \emph{learn} representations by extracting patterns from data, analogous to the formation and refinement of schemas. 

Both schemas and World Models dynamically adapt to new inputs, serving as evolving knowledge representations. However, most machine learning models rely on simple, associative, non-conceptual schemas, lacking the structured abstraction needed for mature reasoning in later developmental stages.

\noindent\paragraph{\textbf{Stages of Learning and Cognitive Development}} As mentioned in the introduction, Piaget’s developmental framework describes four stages of cognitive growth: sensorimotor stage, preoperational stage, concrete operational stage and formal operational stage. In this last abstract, reasoning and hypothetical thinking become possible. 
\begin{tcolorbox}
A World Model can be learned by following a \emph{Piagetian}
cognitive developmental process:
\begin{itemize}[leftmargin=*]
\item A \emph{perception stage}, by which the World Model learns to encode basic input features (e.g., pixel-level data in visual tasks).
\item A \emph{representation stage}, in which the World Model maps raw inputs to structured, semantically interpretable representations (e.g., concepts, objects, relationships).
\item A \emph{reasoning stage}, in which the World Model infers logic between the representations, enabling the model to perform task-specific reasoning.
\item A \emph{generalization stage}, by which the World Model extrapolates such logic reasoning to novel, unseen scenarios.
\end{itemize}
\end{tcolorbox}

Despite the conceptual alignment between these stages and Piaget’s developmental framework, current AI models do not explicitly implement this structured progression. Most modern AI systems, particularly LLMs and deep learning models, operate at the perception and representation stages, excelling in pattern recognition but lacking structured reasoning and generalization. While some architectures incorporate elements of logical inference, their reasoning remains superficial, often emerging as an artifact of learned correlations rather than true causal understanding. Furthermore, AI generalization is limited by training data distributions, whereas human cognition, progressing through Piaget’s stages, develops robust abstraction, enabling flexible adaptation to novel contexts.

\noindent\paragraph{\textbf{{Learning through Interaction}}}

Piaget emphasized that children actively construct knowledge by interacting with their environment, testing hypotheses, and observing outcomes. Children test ideas (like dropping a toy to see what happens) and adjust their understanding based on the results. Machines also learn by interacting with data. In techniques like reinforcement learning, a model tries different actions, observes the outcomes, and adjusts its approach to improve performance. Both systems rely on this trial-and-error process to refine their understanding of the world. The shared reliance on iterative interaction for learning underscores a core functionality of biological systems: they refine their internal representations by engaging with and responding to their surroundings. World Models should operate likewise.

However, current machine learning algorithms fall short of this paradigm. While reinforcement learning incorporates elements of trial-and-error, it often lacks structured abstraction and long-term knowledge accumulation. Unlike human learning, where experiences contribute to a continuously evolving and hierarchical understanding of the world, most models optimize for narrow objectives within a fixed training distribution. This results in representations that fail to generalize beyond specific tasks or environments. Additionally, most methods are passive learners, relying on static datasets rather than active exploration. Children ask questions, experiment, and adapt dynamically, whereas current AI systems remain constrained to patterns that preexist in their training data. Without the ability to autonomously seek information, verify hypotheses, or build cause-and-effect models, AI systems lack the flexible, interactive learning that Piagetian development requires.

\noindent\paragraph{\textbf{{Assimilation, Accommodation, and World Model Updates}}} The dual processes of assimilation and accommodation are pivotal in Piaget’s theory. Assimilation involves integrating new information into existing schemas, while accommodation requires restructuring schemas to incorporate novel or conflicting data. In World Models, these processes should also have direct analogs: in the assimilation phase the model would recognize and fit new data into its existing patterns, whereas in the accommodation phase the World Model would undergo retraining or fine-tuning to adapt to outlier data or changing environments. Both paradigms showcase mechanisms for incorporating new information while preserving coherence within the system’s knowledge base.

However, current machine learning architectures struggle to implement these processes effectively. Most models exhibit a form of rigid assimilation, only absorbing new data within the constraints of their prelearned representations. They lack true accommodation, as they do not dynamically restructure their internal frameworks in response to fundamentally new or conflicting information. Instead, when faced with out-of-distribution data, these systems either fail outright or require computationally expensive full-scale retraining, rather than adjusting knowledge incrementally as humans do.

Furthermore, modern AI systems lack schema-like organization, meaning that new information is often absorbed in an ad hoc manner rather than being systematically integrated into a structured, hierarchical understanding. In contrast, human cognition continuously revises and refines schemas, ensuring that knowledge remains both coherent and adaptable. Without this mechanism, AI models remain vulnerable to catastrophic forgetting, by which newly learned information overwrites previously acquired knowledge. For AI to truly reason and adapt like humans, World Models must be designed to actively restructure their internal representations through continuous interaction with their environment. They must go beyond passive data absorption to develop mechanisms that self-update, reorganize, and reconcile new information with prior knowledge, mirroring Piaget’s assimilation and accommodation processes in cognitive development.

\noindent\paragraph{\textbf{{Biases and the Need for Quality Data}}} Piaget observed that early stages of cognitive development are characterized by biases, such as egocentrism in preoperational thought. These biases gradually diminish as cognitive schemas become more sophisticated through experience and interaction. Similarly, World Models in AI can exhibit biases inherited from their training data, often reflecting statistical imbalances or limitations present in the data sources. Addressing these biases requires exposing models to diverse and representative datasets, employing corrective techniques, or aligning them with external directives to refine their internal representations.

The quality of data plays a fundamental role in World Models. Just as rich, varied experiences accelerate human learning, well-curated, high-quality data can help AI systems generalize effectively while reducing harmful biases. Without this, models risk reinforcing flawed patterns, hindering their ability to reason flexibly and adapt to new situations. These parallels suggest that World Models must overcome these limitations through iterative quality data acquisition and incremental modeling, progressively refining their understanding to achieve robust and unbiased generalization.

\noindent\paragraph{\textbf{Generalization, Abstract Thought and Management of the Unknown}} The culmination of Piaget’s developmental stages is the formal operational stage, where individuals develop the ability to generalize principles and reason abstractly, such as understanding gravity not just as a ball falling but as a universal force. While modern AI, particularly large-scale models, can generalize across tasks by applying learned patterns, this does not reflect true abstract reasoning. Models do not understand the principles they appear to apply; rather, their responses align with correct reasoning patterns by chance, as a byproduct of statistical regularities in training data. This underscores a critical gap between AI’s statistical generalization and the structured, deliberate reasoning of human cognition.

A major contributing factor to this limitation is the absence of true conceptual learning in modern AI. Human cognition organizes knowledge into hierarchies, enabling abstraction, transfer, and compositional reasoning. In contrast, deep learning models rely on statistical associations without structured world representations. This hinders out-of-distribution generalization, causal inference, and symbolic reasoning—core aspects of Piagetian cognitive development. As a result, AI extrapolates surface patterns rather than engaging in deep, principled reasoning. The ability to transition toward broader, abstract concepts is essential for both human and machine intelligence.

\section{Key Research Areas for Constructing World Models}

Piaget’s cognitive development theory suggests that World Models should start with simple building blocks and grow into complex, adaptable frameworks. Current machine learning pipelines are not far away from realizing this vision. However, the predominant approach by which large models are trained on vast datasets and subsequently aligned through post-hoc interventions stands in contrast to Piaget’s developmental theory. Instead of gradually constructing structured knowledge through iterative interactions, modern AI systems passively absorb statistical correlations and later undergo external adjustments to mitigate biases or improve alignment. This reactive paradigm lacks the intrinsic mechanisms of structured learning, conceptual adaptation, and interactive refinement that characterize human cognitive development.

\begin{figure}[h]
  \centering
  \includegraphics[width=\linewidth]{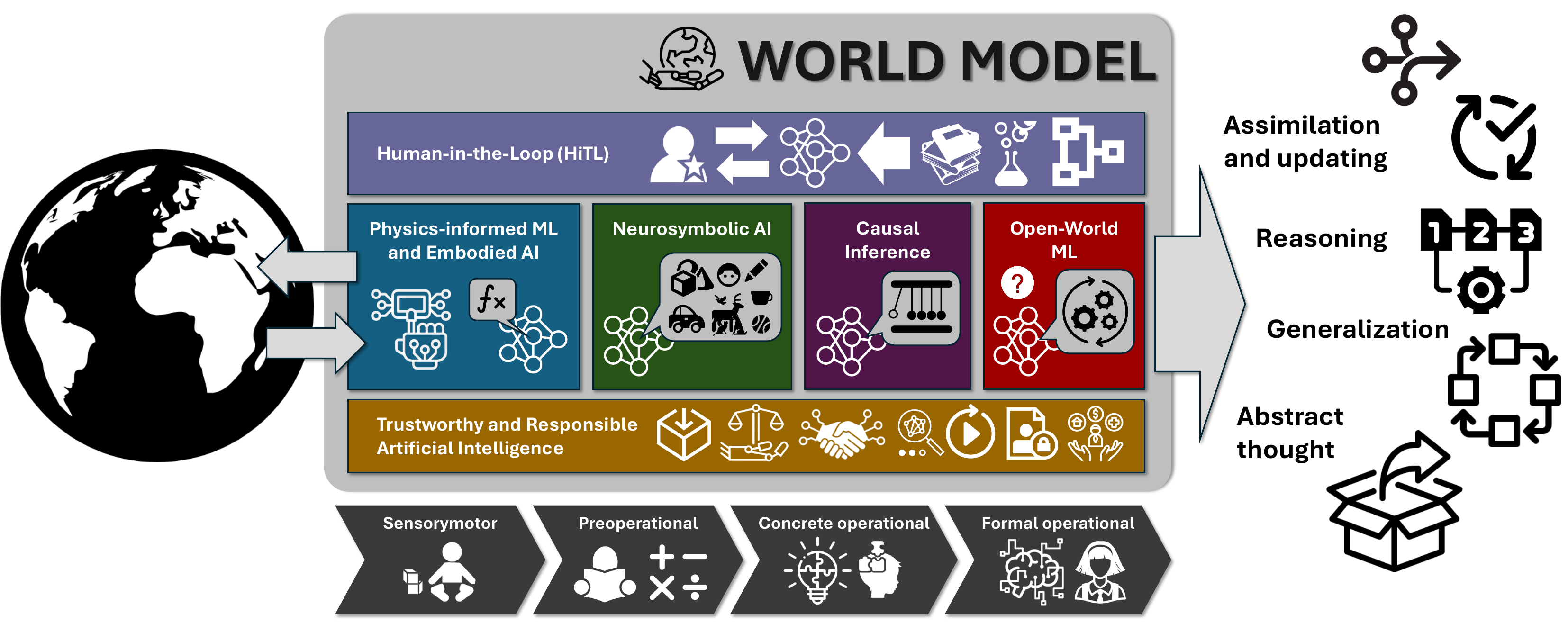} 
  \caption{We envision a World Model built upon 6 research areas: 1) embodied AI and physics-informed machine learning, allowing AI to interact and grasp fundamental relationships about the real world; neuro-symbolic approaches that integrate statistical learning and symbolic representations of the world; 3) causal inference, to capture action-effect mechanisms observed in its collected data; 4) open-world machine learning, to ensure adaptability to unknown situations; (5) Human-in-the-Loop, infusing common sense and oversight; and (6) Trustworthy and Responsible AI, fostering accountability and reliability. Together, these key research area can realize a reasoning-capable, human-centered approach to World Models in AI.}
  \Description[World Model built upon six research areas]{We envision a World Model built upon 6 research areas: 1) embodied AI and physics-informed machine learning, allowing AI to interact and grasp fundamental relationships about the real world; neuro-symbolic approaches that integrate statistical learning and symbolic representations of the world; 3) causal inference, to capture action-effect mechanisms observed in its collected data; 4) open-world machine learning, to ensure adaptability to unknown situations; (5) Human-in-the-Loop, infusing common sense and oversight; and (6) Trustworthy and Responsible AI, fostering accountability and reliability. Together, these key research area can realize a reasoning-capable, human-centered approach to World Models in AI.}
  \label{fig:ourworldmodel}
\end{figure}

To bridge this gap, we highlight six key research areas in AI that are crucial for constructing a Piagetian approach to World Models. These research areas are not isolated but rather interdependent, forming a cohesive framework for building reasoning-capable World Models. To illustrate this interconnectedness, Figure \ref{fig:ourworldmodel} summarizes our perspective on how Piaget’s theory can guide the development of World Models, showing how these research domains interact to enable AI systems that construct, refine, and apply structured knowledge analogously to human learning.

\noindent\paragraph{\textbf{Physics-Informed Machine Learning and Embodied Artificial Intelligence}}Physics-Informed Machine Learning (PIML) enhances traditional AI by integrating physical laws into the learning process, improving generalizability and reducing dependence on large datasets \cite{Karniadakis:2021:PIML}. Unlike purely data-driven approaches, PIML embeds structured knowledge, allowing models to infer fundamental schemas such as physical laws and partial differential equations (PDEs). This is particularly relevant to Piaget’s framework, where learning involves constructing abstract schemas from experience. By incorporating causal constraints, PIML helps AI develop more robust World Models and reduces its dependence on extensive datasets, paralleling how children generalize from limited observations.  

On the other hand, embodied AI extends this by situating learning within the context of an agent’s physical interactions with the world \cite{duan2022survey}. Unlike disembodied models trained on static datasets, embodied AI systems continuously refines its World Model through perception and action, mirroring Piagetian developmental learning. Sensorimotor experiences provide grounded, multimodal data, reinforcing causal relationships and enabling adaptive generalization. This is critical for learning abstract principles, such as object permanence and intuitive physics, through direct interaction rather than passive observation. By combining PIML’s structured knowledge embedding with embodied AI’s experiential learning, AI systems can develop richer, more human-like World Models that evolve dynamically in response to real-world interactions and construct abstract schemas from sensory experiences and interactions.

\noindent\paragraph{\textbf{{Neurosymbolic Systems}}}

The integration of knowledge representation and reasoning into deep learning is not a new idea. Neurosymbolic approaches have been an active area of research for years, aiming to combine the perceptual strengths of neural networks with the structured reasoning and explainability of symbolic AI \cite{Garcez:2023:NeSY}. This fusion allows overcoming the limitations of statistical learning by endowing AI with structured knowledge, logical inference, and interpretability.  

At its core, neurosymbolic AI seeks to bridge two complementary paradigms. Neural networks excel at extracting patterns from raw sensory data, making them particularly effective in handling large-scale, unstructured inputs such as images, speech, and text. However, they lack the ability to explicitly represent abstract concepts, apply structured rules, or perform deductive reasoning. In contrast, symbolic AI (rooted in \emph{Good Old-Fashioned Artificial Intelligence} \cite{McCarthy:1981:AI}) encodes knowledge in formal structures like ontologies, logic rules, and hierarchical representations. By integrating symbolic representations into neural architectures, neurosymbolic systems can acquire the ability to manipulate structured concepts, infer new knowledge, and generalize robustly across domains.  

This integration aligns closely with Piaget’s theory of cognitive development, where intelligence emerges from the interplay between sensory-motor experiences and conceptual reasoning. Just as children first perceive the world through raw stimuli before structuring knowledge into schemas, neurosymbolic AI can provide a framework for machines to transition from passive pattern recognition to active reasoning. By embedding symbolic structures into neural models, AI can refine its internal representations through logical constraints, contextual feedback, and structured adaptation—mirroring how human cognition evolves through assimilation and accommodation.  

Ultimately, neurosymbolic systems hold intrinsic value in constructing more reliable, interpretable, and cognitively grounded World Models. They allows not only recognizing and categorizing stimuli, but also reasoning about their implications, apply learned principles, and adapt to novel situations in a structured manner. This hybrid approach is a critical step toward AI systems that do not simply correlate data, but understand and interact with the world in a meaningful way.

\noindent\paragraph{\textbf{Causal Inference}} Causal inference in machine learning refers to the process of identifying and modeling causal relationships between variables, rather than just detecting correlations \cite{pearl:09,nogueira2022methods}. It allows a model to understand not only how variables are related but also how changes in one variable might directly cause changes in another. In Piaget’s cognitive framework, causal inference is closely tied to the development of logical reasoning, particularly in the concrete and formal operational stages, where children begin to understand cause-and-effect relationships in a structured manner. The ability to grasp these relationships is a significant milestone in cognitive development and is foundational for making predictions and reasoning about the world.

Causal inference is critical for World Models because it enables them to predict the outcomes of actions, understand the consequences of different scenarios, and build robust, causal representations of the environment. In Piaget’s framework, children build causal schemas by interacting with their world, gradually learning how actions lead to predictable outcomes. For AI, causal inference helps World Models reason about underlying mechanisms among the concepts they sense from the worlds. This capability is essential for solving complex tasks that require understanding the \emph{why} behind the observed patterns, and adjusting decision-making based on causal understanding rather than statistical associations.

\noindent\paragraph{\textbf{Open-World Machine Learning}} Open-World Machine Learning (OWML) \cite{kejriwal2024challenges,parmar2023open} refers to AI systems designed to operate in dynamic, unpredictable environments where new concepts, tasks, or data distributions continuously emerge. Unlike traditional closed-World Models, which assume a fixed set of categories and well-defined data distributions, OWML enables AI to identify and adapt to novel and unknown situations without exhaustive retraining \cite{BARCINABLANCO2024128073}. Key subareas include novelty detection (identifying unfamiliar patterns or anomalies), unknown rejection (avoiding confident predictions on unfamiliar inputs), and continual learning (CL), also known as lifelong learning, which allows models to adapt to new environments, tasks, or data over time while retaining previously learned knowledge. CL ensures that World Models can evolve while avoiding catastrophic forgetting, enabling flexible context-aware reasoning.

World models under Piaget’s framework must support accommodation, i.e. the restructuring of cognitive schemas to integrate new information. CL is essential for World Models because it addresses the challenge of incorporating new, sometimes conflicting, information while preserving past knowledge \cite{wang2024comprehensive}. Just as children refine their understanding through experience (progressing from simple sensory-motor interactions to abstract reasoning in the formal operational stage), World Models must continually refine their representations to handle increasingly complex and abstract concepts. By dynamically adjusting without discarding previously learned knowledge, greater coherence and adaptability can be achieved, ensuring that World Models can generalize effectively across a wide range of real-world scenarios.

\noindent\paragraph{\textbf{Knowledge Injection and Human-in-the-Loop (HiTL)}} HiTL in AI integrates human expertise at key stages of the pipeline, enabling iterative feedback, correction of model predictions, and incorporation of domain knowledge to mitigate bias and enhance generalization \cite{mosqueira2023human}. Modern approaches use active learning, where models query humans for the most informative data points, optimizing learning with minimal effort. Knowledge injection embeds explicit knowledge (such as ontologies, rules, and constraints) directly into algorithms, ensuring alignment to domain principles and improving reliability in complex tasks. These methods are widely used in fields like healthcare, where expert input is crucial, and natural language modeling tasks, where human feedback enhances contextual understanding \cite{zanzotto2019human}. 

Jean Piaget's cognitive development theory posits that children construct understanding through active engagement, guided by adults. Similarly, in HiTL AI, human experts reinforce cause-effect relationships, correct errors, and shape learning through feedback \cite{Retzlaff:2024:JAIR}. Just as parents guide children by refining their misconceptions and reinforcing norms through repeated experiences, humans can help AI correct errors, improving causal reasoning and alignment with desired outcomes. This dynamic, iterative interaction between humans and AI models mirrors how children internalize knowledge under the Piaget's theory, positioning HiTL as a crucial area for developing AI with deeper understanding and adaptive learning guided by human supervision.

\noindent\paragraph{\textbf{Trustworthy AI Tools and Responsible AI}} Trustworthy AI tools are designed to ensure that AI systems operate in ways that are reliable, fair, and aligned with human values. These tools focus on enhancing the accountability, robustness, and transparency of AI systems, ensuring that their actions and decisions can be trusted by users \cite{diaz2023connecting}. In Piaget's theory, as cognitive development progresses, children develop an understanding of trustworthiness and the capacity to assess the reliability of their experiences and the people around them. Similarly, AI systems, particularly World Models, need mechanisms that allow humans to trust their reasoning and decision-making capabilities. This is crucial for their safe and effective integration into complex, high-stakes environments where decisions have to be safe, and human oversight is hence necessary.

For World Models, trustworthy AI tools are essential in making sure that their reasoning processes are aligned with human values and expectations \cite{ji2023ai}. Piaget’s framework highlights the importance of developing cognitive structures that can be trusted to navigate and understand the world. In the case of AI, trustworthiness goes beyond the accuracy of predictions; it extends to the transparency, fairness, and safety of the system. By incorporating trustworthy AI tools, World Models can be made more robust, reliable, and dependable, ensuring that their reasoning is not only technically sound but also ethically responsible. 

In this regard, responsible AI frameworks focus on the ethical deployment of AI technologies, ensuring that AI systems operate in a manner that is consistent with societal norms, legal standards, and ethical guidelines \cite{dignum2019responsible,cheng2021socially}. In Piaget's theory, the development of ethical understanding is a key aspect of cognitive growth, as children learn to respect social norms and embrace moral frameworks. Similarly, for World Models, implementing responsible AI frameworks is critical to ensuring that the decisions made by these models adhere to ethical principles and societal expectations, ensuring that AI systems do not cause harm and operate in a manner that aligns with human values, and ultimately fostering societal trust.

\section{Challenges and Outlook: We need World Models}

The development of World Models grounded in Piaget’s cognitive framework represents a significant step toward AI systems capable of reasoning, adaptation, and structured learning. We have identified six key research areas—physics-informed ML, neurosymbolic learning, open-world machine learning, causal inference, human-in-the-AI-loop, and trustworthy and responsible AI—that are essential to constructing these models. By integrating statistical learning with structured reasoning and human expertise, World Models can evolve beyond mere pattern recognition, gaining the ability to form conceptual representations, engage in causal reasoning, and dynamically refine their understanding of the world.  

Despite this promising direction, several challenges remain. The complexity of hybrid models, balancing statistical learning with symbolic reasoning, poses difficulties in interpretability, consistency, and scalability. Ensuring that World Models remain transparent and explainable is crucial for fostering trust, particularly in high-stakes applications. Additionally, discrepancies between learned statistical patterns and predefined logical rules can create conflicts, requiring sophisticated integration methods. The role of human expertise is central to overcoming these hurdles, both for refining models iteratively and for ensuring that they align with ethical and regulatory standards.  

While the path to cognitively inspired World Models is still unfolding, it is important to recognize the significant contributions of large-scale machine learning models in the current AI landscape. These models have proven highly effective in tasks that rely on associative learning, such as language understanding, pattern recognition, and multimodal data processing. However, for AI systems to achieve deep reasoning capabilities, structured world representations, and meaningful generalization, a paradigm shift toward Piagetian-inspired learning is necessary. The future of AI lies in bridging these approaches, leveraging the strengths of large-scale models while incorporating mechanisms for structured learning, reasoning, and continuous adaptation.
\begin{tcolorbox}
\textbf{Takeaway.} Bridging Piaget's constructivist view to human learning with the modeling capabilities of modern AI approaches is a realistic and feasible path towards Artificial Intelligence systems capable of true reasoning, adaptation, and effective generalization.
\end{tcolorbox}


\begin{acks}
Javier Del Ser and Jesus L. Lobo acknowledge funding support from the Basque Government through the ELKARTEK program (BEREZ-IA project, KK-2023/00012) and the consolidated research group MATHMODE (ref. T1456-22). Andreas Holzinger acknowledges funding support from the Austrian Science Fund (FWF), Project: P-32554 explainable Artificial Intelligence. 
\end{acks}

\bibliographystyle{ACM-Reference-Format}
\bibliography{references}

\end{document}